# The More Similar, the Better? Associations between Latent Semantic Similarity and Emotional Experiences Differ across Conversation Contexts

Chen-Wei Yu[1], Yun-Shiuan Chuang[2], Alexandros N. Lotsos[1], Tabea Meier[3,4], and Claudia M. Haase[1]

[1] Northwestern University, USA

[2] University of Wisconsin-Madison, USA

[3] Department of Psychology, University of Zurich, Switzerland

[4] Healthy Longevity Center, University of Zurich, Switzerland

**Abstract**

Latent semantic similarity (LSS) is a measure of the similarity of information exchanges in a conversation. Challenging the assumption that higher LSS bears more positive psychological meaning, we propose that this association might depend on the type of conversation people have. On the one hand, *the share-mind perspective* would predict that higher LSS should be associated with more positive emotional experiences across the board. The *broaden-and-build theory*, on the other hand, would predict that higher LSS should be inversely associated with more positive emotional experiences specifically in pleasant conversations. Linear mixed modeling based on conversations among 50 long-term married couples supported the latter prediction. That is, partners experienced greater positive emotions when their overall information exchanges were more dissimilar in pleasant (but not conflict) conversations. This work highlights the importance of context in understanding the emotional correlates of LSS and exemplifies how modern natural language processing tools can be used to evaluate competing theory-driven hypotheses in social psychology.

**The More Similar, the Better? Associations between Latent Semantic Similarity and**

**Emotional Experiences Differ across Conversation Contexts**

People's use of language is often viewed as a verbal representation of their psychological processes (Boyd & Schwartz, 2021; Sap et al., 2022). With advancements in natural language processing (NLP) technology, researchers have become increasingly interested in questions beyond individual's language use and started probing interpersonal language processes (Giles, 2016; Horn & Meier, 2022; Yeomans et al., 2023). For example, studies have examined how language use in dyadic interactions relates to relationship quality and stability (e.g., Borelli et al., 2019; Ireland et al., 2011; for overviews, see Horn & Meier, 2022 and Karan et al. 2019). One question that has piqued the interest of many is the emotional meaning of similarity in people's language use in conversations (also called linguistic synchrony; Babcock et al., 2014; Bierstetel et al., 2020; Bowen et al., 2017; Brinberg & Ram, 2021; Doré & Morris, 2018; Yeomans et al., 2023). In the present paper, we capitalized on recent technological advancements in NLP to investigate associations between latent semantic similarity (LSS) during married couples' conversations and their subjective emotional experiences. In response to calls for more attention to the role of context in the study of language and social psychology (Harwood & Tracy, 2021; see also Giles, 2016), we furthermore examined whether these associations varied across two different conversation contexts—pleasant and conflict conversations.

**Latent Semantic Similarity**

Latent semantic similarity (LSS) is a measure derived from latent semantic analysis. Latent semantic analysis is an NLP technique that captures the latent semantic information of a text, such as a word or a sentence, and represents this information in numeric form, usually as a vector (i.e., embedding; Demszky et al., 2023; Landauer & Dumais, 1997). The similarity between two pieces of latent semantic information is the latent semantic similarity.

### Latent Semantic Similarity and Emotional Experiences in Context

Applied in dyadic studies in social psychology, LSS is thought to capture the mutual attentiveness of the conversation partners, indicating that both parties are engaged in the exchange of information on the same topic (Babcock et al., 2014; Rossignac-Milon et al., 2021). As a result, high LSS between two people is theorized to be an emergent property of a "shared mind" of the dyad (Higgins et al., 2021), indicating that both conversation partners are "on the same page". Both of these metaphors have been theorized to be associated with positive emotional experiences during interpersonal interactions (Rossignac-Milon & Higgins, 2018; Wolf & Tomasello, 2023; also see Reis et al., 2021). From this view, interacting partners whose conversations are characterized by greater LSS would experience more positive emotions, regardless of what they are talking about (see Higgins et al., 2021).

Yet, while it might be good for conversations about conflict resolution or consensus building for both partners to try to stay highly focused on the target issue (Gottman, 2017; Levenson, 2024), in pleasant conversations, such hyperfocus on a single topic might instead suggest a shortage in topics of shared enjoyment or pleasant memories to talk about (Bietti, 2010; Hirst & Echterhoff, 2012; Pasupathi et al., 2002). Hence, the association between LSS and people's emotional experiences during a conversation might depend on the *context* of the conversation (for a brief review on the importance of context in the study of language and social psychology, see Harwood & Tracy, 2021).

To research the emotional correlates of LSS through this contextual lens, it is useful to contrast two conversation contexts that have been widely studied in couples research since the 1980s—conflict and pleasant conversations (Levenson, 2024). For conflict conversations—the most widely studied conversation context (Gottman, 2017; also see Haase, 2023; Horn & Meier, 2022)—couples are prompted to talk about an area of disagreement in their relationship and try to come to a resolution. As a result, it is important for both partners to stay on topic and try to reach a consensus. In this situation, higher LSS might indicate that both partners are "on the same page," and might experience more positive emotional experiences (e.g, Gottman, 2017), as the shared-mind perspective suggests (Rossignac-Milon & Higgins, 2018).

By contrast, in pleasant conversations, where people are prompted to talk about a topic of mutual enjoyment, this association might be reversed. That is, higher LSS might be linked with lower positive emotional experiences in pleasant conversations. We derived this prediction from the *broaden-and-build* theory of positive emotions, which posits that positive emotions are associated with a widened "array of thoughts, action urges, and percepts that spontaneously come to mind" (Fredrickson, 2013; p.17). This suggests that, if people experience positive emotions during a conversation, they would have an expanded cognitive space from which to draw shared pleasant or enjoyable conversation topics (see Pasupathi et al., 2002). With a widened cognitive space, conversation partners may also be able to build off from each other's sharing more easily, guiding the conversation to other related shared pleasant memories (Templeton & Wheatley, 2023). Hence, the conversation topics might diverge in multiple directions instead of converging on one point in the semantic space. Therefore, in pleasant conversations, lower LSS might be associated with greater positive emotional experiences.

**The Present Study**

The present study examined links between latent semantic similarity and subjective emotional experiences across different conversation contexts. Drawing from the shared-mind perspective (Higgins et al., 2021) and the broaden-and-build theory of positive emotions (Fredrickson, 2013), we tested two competing hypotheses (H) regarding the link between LSS and people's emotional experiences within each conversation context:

H1: Higher latent semantic similarity will be associated with greater positive emotional experiences for partners across conversation contexts, according to the shared-mind perspective.

H2: Lower latent semantic similarity will be associated with greater positive emotional experiences for partners in the *pleasant* (but not conflict) conversation, according to the broaden-and-build theory.

We tested these hypotheses using data from a laboratory-based dyadic interaction study with long-term married couples who had been married for 11 years on average by the time of the study. Couples engaged in two naturalistic 10-minute conversations about a topic of disagreement (i.e., conflict conversation) and a topic of mutual enjoyment (i.e., pleasant conversation) and reported on the valence of their subjective emotional experiences after each conversation. The nature of long-term marriage provided a theoretically appropriate means for our empirical examination. Married couples have both important longstanding relationship conflicts as well as topics of mutual enjoyment to talk about (see Levenson, 2024). In the case of strangers, acquaintances, or even newly made friends, people may not have a long enough history to either have conflicts or share pleasant memories (Planalp, 1993; Vieth et al., 2022). Hence, lower LSS in these dyads might reflect individuals' difficulty in maintaining the conversation more than the dyad's shared experiences (Planalp, 1993; Templeton & Wheatley, 2023).

We quantified latent semantic similarity using sentence embeddings where individual input sentences are transformed into fixed-size, high-dimensional numeric vectors (i.e.

embeddings; for reviews on text embedding and applications in psychology, see Boyd & Schwartz, 2021 and Jackson et al., 2021). Sentence embeddings preserve the contextualized semantic meaning of the input sentences by considering the meaning of each word within the context of the whole sentence when constructing the embedding. This means that the same word can have different vector representations depending on how it is used in a sentence (e.g. "bank" in "river bank" vs "financial bank"), thus presenting a context-aware representation that helps capture the linguistic subtleties present in natural interactions (R. Li et al., 2022; Reimers & Gurevych, 2019). Because this approach has not been adopted before in dyadic studies to our knowledge, we carefully devised three tests to validate our measure (see Method for detail; Baden et al., 2022). Finally, with the validated LSS, we employed linear-mixed modeling to statistically evaluate our hypotheses (Kenny et al., 2020).

We checked the robustness of our results by evaluating whether findings remained stable when controlling for (a) positive and negative emotional word use (Meier et al., 2024) computed from *Linguistic Inquiry and Word Count* (LIWC) software (Boyd et al., 2022) (b) emotional language style matching (Doré & Morris, 2018), (c) linguistic style matching (Niederhoffer & Pennebaker, 2002), (d) the number of utterance pairs, and (e) marital satisfaction.

We controlled for (a) to ensure that our effect estimate was beyond each individual's use of *positive* and *negative emotional words*, which have been found to be associated with people's emotional experiences (Meier, Stephens, et al., 2024; Yu & Chang, 2023). In terms of (b), which was conceptually defined as the similar use of positive and negative emotional words in two people's language (see Method section for more detail; also see Doré & Morris, 2018), we planned to rule out the explanation that our effect of interest was driven by similarity in emotional words instead of in semantic meanings. We controlled for (c), which was similarly defined as (b) but in terms of the use of *function words*, to rule out the explanation that our effect of interest was driven by the (theoretically) automatic cognitive process of the use of function words (Bierstetel et al., 2020; Niederhoffer & Pennebaker, 2002). We controlled for (d) and (e) because they were both potential omitted variables that could be associated with both overall semantic similarity and people's emotional experiences.

## Method

### Participants and Procedure

Fifty-seven married mixed-sex[1] couples (114 individuals) from diverse racial and socioeconomic backgrounds were recruited in the Chicago metropolitan area for a larger research project on socioemotional functioning in married spouses who had at least one child aged 5 to 18 years. Findings from this project have been published (Hittner et al., 2019; Hittner & Haase, 2021; Meier, Otero, et al., 2024; Meier, Stephens, et al., 2024), but no previous studies have examined associations between LSS and emotional experiences.

We recruited participants through flyers, advertisements on public transportation, and online postings between 2015 and 2017. Participants were invited to the laboratory and engaged in two audio- and video-recorded 10-minute naturalistic conversations as they would in their home or in a diner (Gottman & Gottman, 2017; Levenson, 2024; Levenson & Gottman, 1983). These conversations were about (a) a topic of mutual enjoyment (hereafter pleasant conversation), and (b) a topic of disagreement with the instruction to try to come to a resolution (hereafter conflict conversation), order counterbalanced across couples. After each conversation, participants were asked to "*indicate how positive or negative you felt during the conversation you just had*" on a scale ranging from 1 = *Very unpleasant* to 9 = *Very pleasant*. They also reported their perceived naturalness of the conversation on the item "*How similar was this conversation to conversations you have in daily life?*" on a scale ranged from 0 = *Not at all similar* to 8 = *Very similar*. We used this item as an ecological validity check. Finally, after all the conversations, participants completed questionnaires measuring marital satisfaction (Locke & Wallace, 1959), sociodemographic characteristics, and other variables unrelated to the present study. The study lasted for approximately three hours and participants were each compensated with $100.

---

[1] Our sample consisted of mixed-sex married couples due to local marriage legislation when data were collected.

Data were missing for eight couples' conversations (seven from pleasant and eight from conflict conversations) due to cameras or microphones malfunctioning and for two individuals' emotion ratings (from both conversations) due to iPads malfunctioning (used to fill out survey), resulting in 50 and 49 couples in the analyses for pleasant and conflict conversations, respectively.[2] The final sample (of 49 couples) were 42.20% white, 36.70 % African American/Black, 7.34 % Latinx/Hispanic, 7.34 % Asian/South Asian, and 6.42% identified as other racial-ethic categories. As for socioeconomic status, there were 14.3% of participants with an income (defined as annual household income before taxes) less than $20,000, 12.2% with an income between $20,001–$35,000, 14.3% with an income between $35,001–$50,000, 12.2% with an income between $50,001–$75,000, 14.3% with an income between $75,001–$100,000, 20.4% with an income between $100,001–$150,000, and 12.2% with an income greater than $150,000. On average, participants were 42.70 years old (SD = 9.73) and had been married for $M$ = 11.69 years ($SD$ = 6.63) at the time of the study.

We conducted a sensitivity analysis to evaluate the minimum effect size that could be detected based on our sample size at significance level of .05 with 80% statistical power (Giner-

---

[2] Participants with missing data did not differ from others in terms of demographic characteristics, i.e., age, $t(33.2)$ = 1.51, $p$ = .141 and income, $t(49.9)$ = 1.85, $p$ = .070.

Sorolla et al., 2024). Following the procedure of Ackerman and Kenny (2016), we found that our sample allowed us to detect medium-sized effects of $r = .27$ and higher.

**Analytic Strategies**

*Latent Semantic Similarity*

      **Data Preprocessing.** We computed LSS scores based on the word-by-word transcripts of each conversation. We defined a turn as everything a speaker said before they were interrupted by their spouse, following Yeomans et al. (2023). This could be one or multiple sentences. To preserve the semantic meaning of each turn, stop words were not removed (e.g., "I", "maybe"). However, turns consisting of only filler words (e.g., "Um", "Mhm") were removed as they did not contain much semantic meaning and would introduce noise in our similarity measure (see below).

      **Turn Embedding and Semantic Similarity**. Turn embeddings were derived from the sentence transformer *General Text Embeddings-Large* (gte-large) model (Z. Li et al., 2023)[3]. For any turn, the model computed a turn embedding ($e_i$), which was a 1024-dimensional vector capturing the semantic meaning of the turn. For any two temporally adjacent turns $e_i$ and $e_{i+1}$, pairwise LSS ($s$) was obtained by cosine similarity $\cos(e_i, e_{i+1})$, a standard metric to compute the similarity of two embeddings (for a review, see Chandrasekaran & Mago, 2021). For example, if $e_1$, is the embedding of the first turn of a conversation and is produced by the wife, then, by definition, $e_2$ would be the second turn of a conversation and is produced by the husband. For a conversation, we could then compute the overall LSS ($\bar{s}$) between all adjacent turns as follows:

---

[3] At the time of data analysis (August, 2023), this model was the best-performing model in terms of sentence similarity, based on the *Mass Text Embedding Benchmark* (Muennighoff et al., 2023). Our conclusions remained stable even when we used another transformer (see robustness checks).

$$\bar{s} = \frac{\sum_{i=1}^{N-1} \cos(e_i, e_{i+1})}{N-1}$$

where $N$ was the number of turns in the entire conversation. A higher value of $\bar{s}$ would indicate a higher overall LSS of the couple (compared to other couples in the sample).

Importantly, our similarity measures ($s$ and $\bar{s}$) must not be interpreted as *Pearson's correlation coefficient* ($r$). While both are measures of association and can range from -1 to 1, they bear distinct mathematical and conceptual meanings. Moreover, for the model we used in the present paper (i.e., gte-large), cosine similarity always fell in the range between .7 to .1, because the model adapted a contrastive loss function (for technical details, see Li et al., 2023). As a result, it is important to note that the absolute value of our measure cannot be interpreted as effect size (i.e., suitable for cross-sample comparison), and it can only be interpreted in a rank-order fashion (i.e., within-sample comparison). To provide an intuitive sense of what LSS captures, we provide some pairs of conversation turns and their pairwise LSS in Figure 1.

**Validity Checks.** Because there was limited understanding in how sentence embeddings and measures of LSS performed in a dyadic conversation context, we first aimed to establish whether our approach was able to properly consider two important features within this context: (a) temporal continuity and (b) dyadic interdependence (see Wood et al., 2021). We therefore conducted three sets of analyses to ensure that our LSS measure was successful in capturing these features instead of merely collecting noise in the data and that our empirical results were not merely a statistical artifact (Baden et al., 2022). Specifically, the first two sets of analyses addressed the temporal continuity of the data, and the last one addressed the dyadic interdependence thereof.

**Figure 1**

*An example of the trend of pairwise similarity within a pleasant conversation*

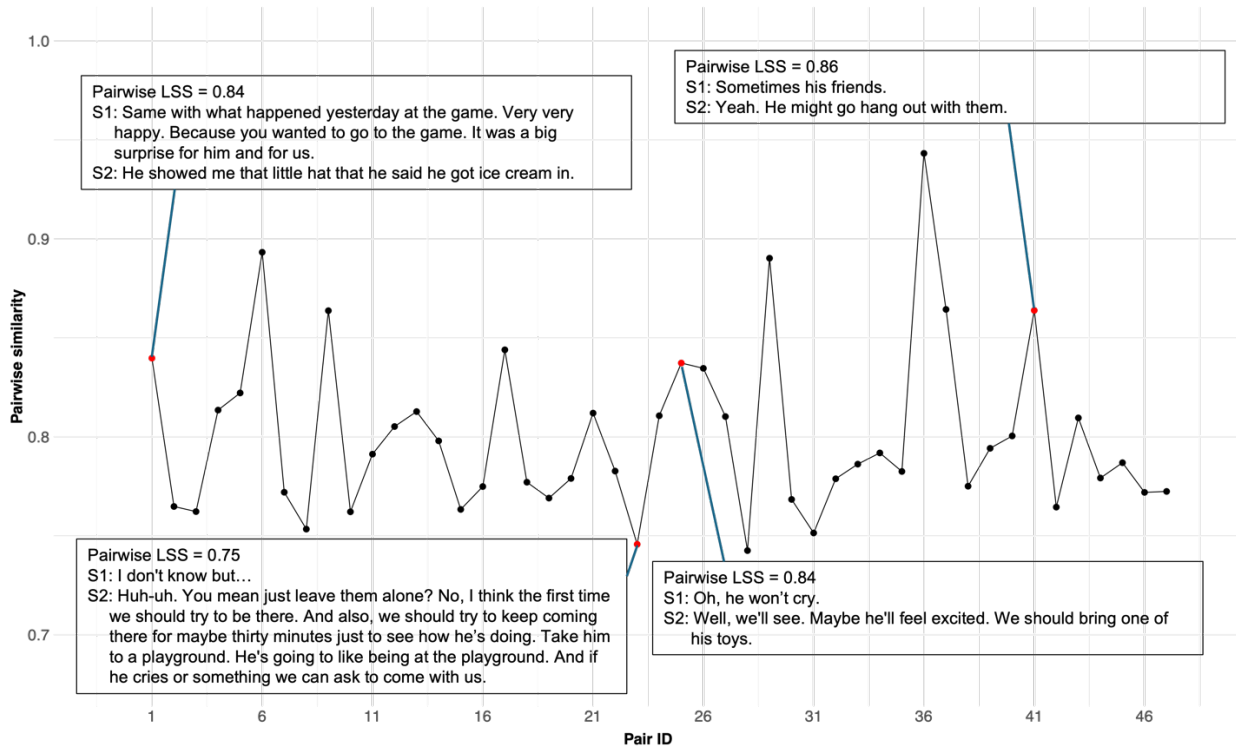

*Note.* To ensure the anonymity of our participants and to protect their privacy, we changed some words in the sentences. S = spouse.

**Validation 1: Similarity between the First and Rest of the Turns.** An intuitive way to establish the importance of the temporal order of the turns was to examine whether the similarity between $e_1$ and $e_i$ decreases as $i$ increases. A decreasing trend would support the validity of our approach because temporally closer turns should have higher similarity compared to farther ones.

**Validation 2: Pseudo Turn Pairs within a Couple.** We employed a permutation test to further address the importance of the temporal order of the turns. For each conversation of a couple, we shuffled the order of the wife's and the husband's turns. Overall LSS ($\bar{s}$) was then computed for each shuffled set. This procedure was repeated 1,000 times, generating a permutation distribution of overall LSS (i.e., a distribution of $\bar{s}$) for each couple and for a conversation type. We then compared the observed value against this distribution and calculated the $p$ value by dividing the number of cases in the distribution that are equal to or larger than the distribution. Because the turn pairs consisting of the distribution did not come from temporally adjacent turns, we expected to find that the observed similarity score to be significantly larger than the permutated similarity score.

**_Validation 3: Pseudo Speaker Pairs across Couples._** The third validation probed the importance of semantic interdependency for a couple engaging in a conversation. To this end, we paired each speaker with another speaker from all the other couples to form a mixed-sex pseudo couple (see Wood et al., 2021). For instance, husband 1 ($H_1$), whose original partner was wife 1 ($W_1$), was paired with each of the wives (i.e., $W_2$, $W_3$, $W_4$, … $W_n$) from all the other couples in the sample. Pairwise LSS ($s$) was then computed based on temporally adjacent turns of the pseudo couples, which average to overall LSS ($\bar{s}$). With this procedure, the turn pairs consisted of turns coming from people not in a relationship and talking about different topics, and therefore the similarity scores for these pseudo couples were expected to be lower than similarity for the real couples. For every real dyad, we conducted a one-sample $t$-test to compare the observed similarity of the real dyad with the distribution of the similarity scores of the pseudo couples.

### Hypothesis Testing

We employed linear mixed-effect modeling to test our hypotheses, which accounted for interdependence between husbands and wives (Kenny et al., 2020). In the model, we entered overall LSS and participant's sex as predictors of participant's emotional experiences. We also included conversation type as a repeated-measure variable following previous research (West, 2013; see also Meier et al., 2024). These variables were allowed to interact with one another. We were specifically interested in the interaction between LSS and conversation type which indicated whether the effect of LSS on emotional experiences varied across conversation types. If the interaction term was significant, we proceeded to analyze two conversations separately to estimate the effects of LSS on emotional experiences in each of the conversation. Here, significant and positive effects of LSS in both conversations would support the *shared-minds* perspective, while a significant and negative effect in the pleasant, but not in the conflict, conversation would support the *broaden-and-build* perspective. All the variables were standardized in these analyses.

### *Robustness Checks*

In five different models, we entered one of the following set of variables as a covariate to estimate the net effect of overall similarity: (a) positive and negative emotional word use computed from the LIWC software (Boyd et al., 2022) (b) emotional language style matching, (c) linguistic style matching, (d) the number of utterance pairs, and (e) marital satisfaction. These variables were respectively entered in separate models instead of simultaneously entered in one model to avoid over-fitting. In a sixth model, we included all the covariates simultaneously.

As a software developed for automated text processing, LIWC took a dictionary approach to labeling the words being used in a text. The labels were validated extensively for their psychological meanings (Boyd et al., 2022; Chung & Pennebaker, 2018). To get a reliable index from LIWC, at least 50–100 words are needed (Boyd et al., 2022). Moreover, because we were interested in conversation-level analysis, we collapsed all the turns from husbands into one text and did the same for wives. The dictionary approach of LIWC means that it is acontextual, and thus this procedure would not affect the labels. We then followed the established procedure of Doré and Morris (2018) to compute emotional language style matching and linguistic style matching. Specifically, these measures were defined in the following form:

$$synchrony_{label} = 1 - \frac{|labels_H - labels_W|}{labels_H + labels_W + 0.0001}$$

where label is either emotional words or propositions, H is the husband, and W is the wife.

Finally, to ensure that our results were not dependent on a specific machine-learning algorithm, we computed turn embedding using another transformer and reran the statistical analyses with the output from this model. We chose the Sentence BERT (SBERT) as an alternative model to derive turn embeddings, which was based on the BERT model (Devlin et al., 2019; Reimers & Gurevych, 2019), a model familiar to many researchers, including psychologists (Biggiogera et al., 2021; Kjell et al., 2023; Rogers et al., 2020).

Overall, if we consistently found our primary effect of interest across these analyses, it would provide strong evidence supporting our conclusion.

## Results

**Ecological Validity and Manipulation Checks**

We first examined whether participants had naturalistic conversations as instructed. To this end, we focused on participants' response to the question assessing their perceived naturalness of the conversation. We compared their response to the scale mid-point (i.e., 4) using one-sample $t$-test. Results showed that participants thought both the pleasant ($M = 7.36$, $SD = 1.77$) and conflict ($M = 7.51$, $SD = 1.59$) conversations were highly similar to their natural conversations, both $t$s > 18, $p$s < .001.

Next, we examined whether participants did have two different kinds of conversations, and whether these conversations elicited the expected emotional experiences in participants. As such, we tested whether participants' emotional experiences were indeed more positive in the pleasant compared to the conflict conversation. A linear mixed model confirmed that conversation context had a significant effect on participants' emotional experience, $b = 1.36$, $p < .001$, and participants, regardless of sex, felt more positive in the pleasant ($M = 7.20$, $SD = 1.62$), compared to the conflict ($M = 5.83$, $SD = 2.04$) conversation.

**Validity Checks**

We first present evidence supporting the validity of our similarity measure before presenting results of our primary hypothesis test.

***Validation 1: Similarity between the First and Rest of the Turns***

First, we examined temporal trends in LSS scores between the first and rest of the turns. As shown in panel a of Figure 2, there was a decreasing trend in the similarity scores as the turns used to compute the similarity became more distant. In other words, the similarity between the first turn and the subsequent turns decreased. This was observed for both the pleasant and the conflict conversations. We only present the trend for the first ten turns, because the values seemed to reach a stable low point, suggesting that turns did not become more and more dissimilar. This analysis provided an intuitive first way to validate that our similarity measure captured the temporal adjacency of turn pairs.

**Figure 2**

*Validity checks of the similarity measure*

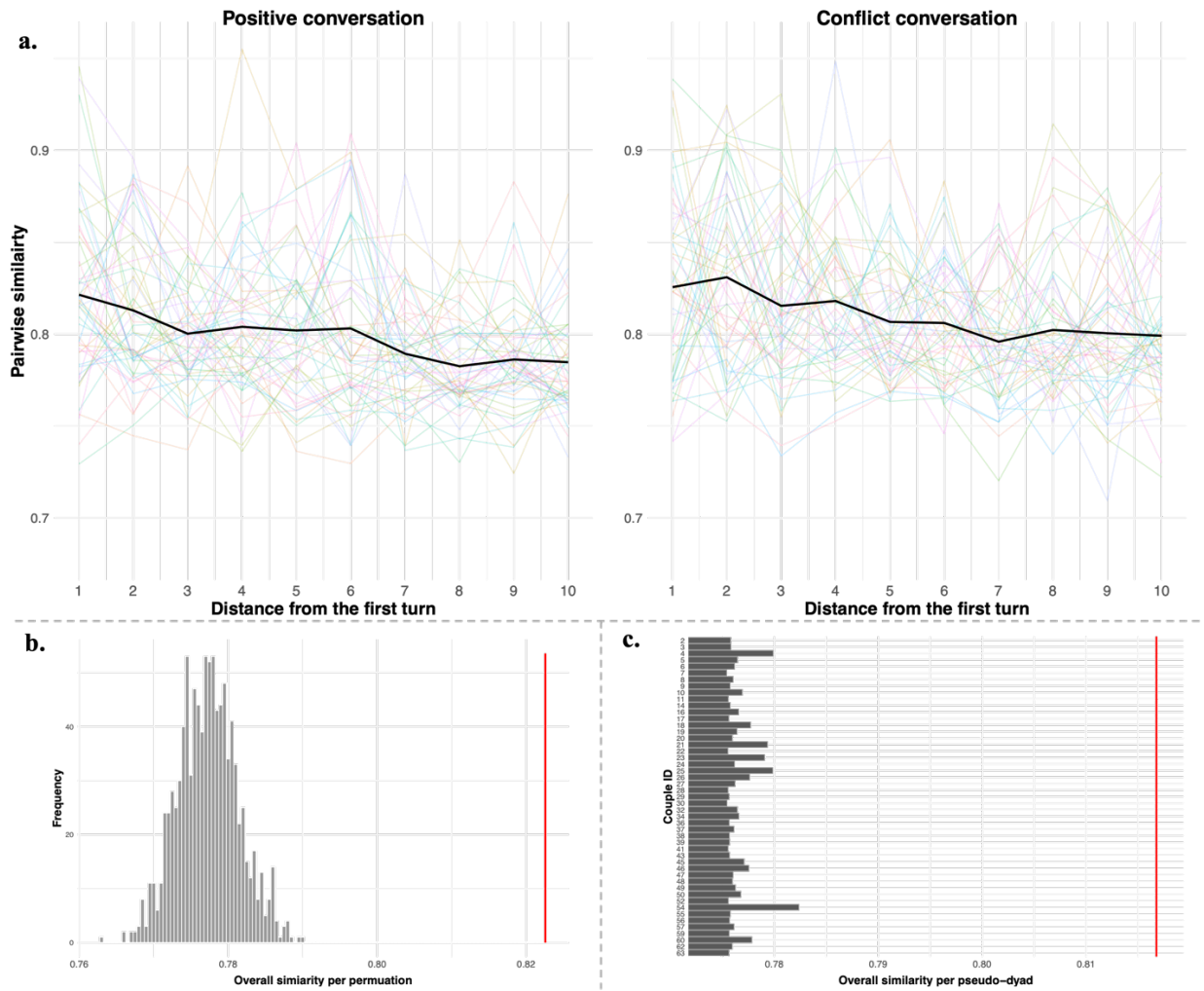

*Note.* Panel a demonstrates the temporal decay of pairwise similarity as the turns used to compute the similarity become more and more distant. The black thick line is the average trend, and the colored thin lines are the trend for each couple. Panel b is the distribution of the overall similarity scores of a couple's 1,000 times permutated turn order. The gray histogram is the distribution of the permuted scores, and the red line is the observed overall similarity for the couple. The observed value shows no indication to be drawn from the randomly ordered turn sample. Panel c demonstrates the comparison between a couple's overall similarity score and

those of pseudo-dyads'. The gray bars are the scores computed based on pseudo-dyads and the red line is the observed score for the real dyad. The observed value shows no indication to be obtained from a pseudo-dyad.

### Validation 2: Pseudo Turn Pairs within a Couple

Second, we examined pseudo turn pairs within a couple. For the pleasant conversation, the observed similarity scores were larger than all 1,000 permutation samples for every couple, $ps < .001$. In other words, when the similarity was computed based on a shuffled set of turns, it did not produce a value comparable to what we observed in the data. The panel b of Figure 2 shows an example from a couple. The figure shows the distribution of the permutated scores (gray histogram), and the red line is the observed overall similarity for the couple. As shown by the distance between the red line and the distribution, the observed value shows no indication to be drawn from the randomly ordered turn sample.

For the conflict conversation, out of the 49 couples, 41 showed observed similarity scores significantly larger than all the permutation samples, $ps < .001$. For the rest, three showed overall similarity scores larger than the permutation samples at $p = .01$, two larger than the permutation samples at $p = .05$, and the last three did not pass the permutation test, $ps > .10$. We suspected that these results could be due to a lower number of turn pairs in these couples, such that our permutation produced similar turn pairs order to the original turn pairs order. We conducted a $t$-test and found support for this hypothesis: There were more turn pairs in couples whose observed overall similarity scores were larger than all 1,000 permutation samples ($M = 77.4$, $SD = 41.5$), compared to those eight couples whose observed overall similarity score was not ($M = 29.8$, $SD$ = 14.4), $t(32.9)$[4] = 5.77, $p < .001$.

These results indicated that our similarity measure captured the temporal adjacency of turn pairs, instead of random pattern in the data.

### Validation 3: Pseudo Speaker Pairs Across Couples

Third, we examined pseudo speaker pairs across couples. As expected, for both conversation types, overall similarity scores for the pseudo couples were all lower than the observed overall similarity scores for the real dyads, $p$s < .001. In other words, when a speaker was paired with another speaker other than their spouse (i.e., forming a pseudo couple), their turns did not produce a similarity value comparable to what we observed in the data. Panel c of Figure 2 shows an example of a dyad. The gray bars are the scores computed based on pseudo-couples and the red line is the observed score for the real dyad. As shown by the distance between the red line and the bars, the observed value based on the real dyad is always larger than the simulated values based on pseudo couples.

This analysis showed that our similarity measure captured the dyadic interdependence of the data, such that speaker turns within a specific conversation of a couple could only be examined within the couple, but not across other couples having similar-valenced conversations (i.e., pleasant/conflict conversations).

Taken together, these validation tests showed that our observed similarity scores were able to capture the temporal continuity (validation tests 1 and 2) and dyadic interdependence (validation test 3) of our conversation data. We then proceeded to statistical analyses with this measure.

---

[4] Degrees of freedom are adjusted with Welch's procedure to adjust for unequal variances across groups.

**Hypothesis Testing**

In the linear mixed-effect model, we found a significant interaction between overall LSS and conversation type on emotional experiences, $\beta$ = -.45, $p$ < .001. We therefore proceeded to examine the effects of overall LSS on emotional experiences separately in each conversation. We present the effect estimates in Table 1.

In the pleasant conversation, we found that overall LSS was a significant negative predictor, $\beta$ = -.30, $p$ < .05, such that lower LSS was associated with greater positive emotional experiences. No significant main or interaction effects were found for sex, $\beta$s < .15, $p$s > .05. By contrast, in the conflict conversation, overall LSS showed a positive, but nonsignificant association with emotional experiences, $\beta$ = .19, $p$ = .262. Again, no main or interaction effects were found for sex, $\beta$s < .10, $p$s > .05.

Figure 3 charts the data points and associations between overall similarity and emotional experiences for each conversation. Figure 1 shows the trend of pairwise similarity of a couple's pleasant conversation. This couple had low overall LSS (percentage rank [PR] among all couples = 4). As can be seen, the couple talked about various topics (e.g., going to a game, taking their kid to the playground and bringing their toy, and their kid hanging out with friends).

**Figure 3**

*Associations between overall semantic similarity and emotional experiences in pleasant and in conflict conversations*

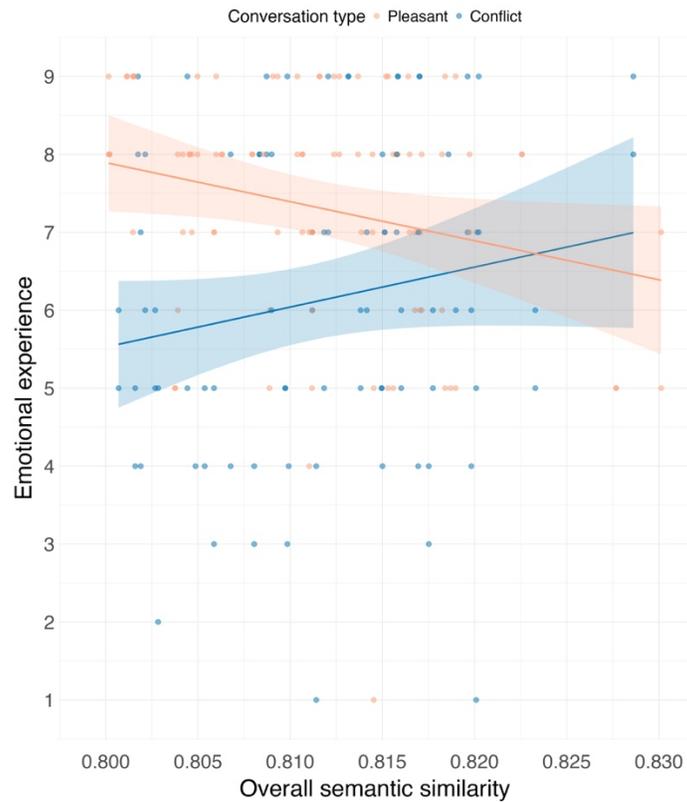

*Note*. Each dot represents the overall semantic similarity of a couple. Shaded areas around the regression lines represent 95% confidence intervals.

**Robustness Checks**

First, we consistently found a significant main effect of overall LSS in the pleasant conversation in all the models with different covariates included, βs ranging from -.28 to -.31, all $p$s < .05. By contrast, there was no significant effect of overall similarity for the conflict conversation, βs ranging from .08 to .24, all $p$s > .10. In supplemental materials, we report the correlations among all the variables as well as the tables for the effect estimates.

Next, we found that the overall LSS based on turn embeddings derived from gte-large and from SBERT were highly correlated, $r = .83$, $p < .011$. We then entered the overall LSS measure from SBERT into linear mixed models to test our hypotheses. Results were very similar to the original findings, except that the effect sizes became slightly smaller. In the pleasant conversation, overall LSS was a negative predictor of emotional experiences with marginal significance, $\beta = -.29$, $p = .061$. In the conflict conversation, overall LSS was a positive, but again non-significant, predictor of emotional experiences, $\beta = .11$, $p = .589$, parallel to our original analyses. Despite the minor drop in effect sizes, the conclusions that could be drawn from the SBERT-based similarity measures remained the same. Taken together, these analyses show that our results are unlikely to change substantially with different choices of algorithms.

## Discussion

Scholars have been deeply interested in the natural emergence and psychological meanings of synchronization in dyadic interactions (Wood et al., 2021) across facial expressions and gestures (Wells et al., 2022), physiological responses (Chen et al., 2020), and the use of language (Babcock et al., 2014; Bierstetel et al., 2020; Doré & Morris, 2018). Challenging the prevalent assumption that higher similarity in people's exchanges of semantic information bears uniformly positive psychological meanings (Rossignac-Milon & Higgins, 2018), the present study revealed context-specific links.

Consistent with the prediction grounded in the broaden-and build theory of positive emotions (Fredrickson, 2013), we found that lower latent semantic similarity was associated with greater positive emotional experiences in the pleasant conversation. In other words, when couples talked about a variety of pleasant topics—in one couple's conversation, a sports game, taking their kid to the playground, and their kid hanging out with friends, for example—both partners ended up feeling more positive. This finding was robust to the inclusion of a variety of theoretically relevant covariates—including the use of emotion words, emotional language style matching, linguistic style matching, the number of utterance pairs, and marital satisfaction. We also found evidence that our results were stable when using a different embedding algorithm.

In contrast, in the conflict conversation, *higher* latent semantic similarity was directionally (albeit not significantly) associated with greater positive emotional experiences. The shared-mind perspective (Higgins et al., 2021) as well as previous marital interaction research (Gottman, 2017) have emphasized the importance of being "on the same page" during conflict conversations, we might add. Bearing in mind that we observed an overall nonsignificant association in this context, future research may explore whether the benefits of greater similarity during conflict conversations may emerge on a cognitive level rather than on an emotional level, reflecting that couples are coordinating and making progress on resolving a disagreement (see Meier et al., 2021, and Horn & Meier, 2022).

Overall, this work underscores the importance of context in understanding the psychological meaning of linguistic synchrony (Gallois et al., 2021), and also exemplifies how modern NLP tools can be used to further explore competing theory-driven hypotheses in social psychology (Harwood & Tracy, 2021).

**Theoretical and Methodological Implications**

The present study contributes to the bourgeoning interdisciplinary interest among affective scientists and relationship scientists in understanding the complex interplay between language use and emotions emerging in natural social interactions (Yeomans et al., 2023). Affective scientists have long sought to understand what shapes people's emotional experiences (e.g., Lench et al., 2011; Shiota et al., 2017). While many studies have linked language use to emotional experiences, these efforts have often been confined to measuring the frequency of specific words, including emotion words (e.g., Meier et al., 2024; Yu & Chang, 2023). Our study suggests that LSS might be a more implicit and contextual correlate of emotional experiences, perhaps less prone to demand characteristics and spontaneously emerging in people's conversations (Horn & Meier, 2022), particularly when involvement is high (Babcock et al., 2014).

Relationship scientists in turn have been interested in what makes for a happy relationship and some have focused in particular on how couples navigate conflict or pleasant interactions (e.g., Feeney & Karantzas, 2017; Gottman, 2017; Harris et al., 2014). Our findings suggest that latent semantic similarity might capture, in part, how well couples navigate different kinds of conversations. When a pleasant interaction is going emotionally well, similarity may be lower. When a conflict is being effectively resolved, the opposite may be the case (though our findings cannot speak to the cognitive evaluation of whether conflict is resolved well, see above). While the correlational nature of our data prevents us from drawing causal conclusion, it is possible that these associations are in fact bi-directional. For instance, couples who experience positive emotions might be able to access and talk about a diverse range of pleasant topics. At the same time, couples who talk about different enjoyable topics may experience a boost in positive emotions. Future research could examine these possibilities and test their applications in important real-world settings, such as couples therapy.

Furthermore, while we grounded our predictions in the shared-mind perspective and the broaden-and-build theory of positive emotions, our finding raises the interesting possibility that LSS might signal a couple's interdependence with each other (Finkel et al., 2017; Rusbult &

Lange, 2003; Wegner et al., 1985)[5]. That is, couples who have high interdependence, because of their highly inter-woven lives, may have many positive experiences together. Therefore, these couples might have a broader space to draw pleasant memories from in the first place (see Harris et al., 2014). When asked to talk about a topic of mutual enjoyment, they may be able to easily access more potential topics and memories than couples low on interdependence. It is also possible that because couples with high interdependence also tend to deal with conflicts more effectively (Feeney & Karantzas, 2017; Meier et al., 2021), these couples might display high LSS in the conflict discussions (see Babcock et al., 2014). Future research can explore the associations between interdependence and LSS derived from different contexts. Converging with other contextualized perspectives on language use in couples (Bowen et al., 2017; Meier et al., 2021), LSS could perhaps be seen as a flexible manifestation of how interdependent couples navigate the highs and lows of their relationship. This an interesting empirical question for future research.

        Finally, the present study has methodological implications. We demonstrated three theory-informed validation tests to ensure the ability of our LSS measure to capture the dyadic nature of our data. These validations can be readily applied by researchers interested both in other ways of measuring LSS and in other quantitative indices composed of two individual's interdependent data. Moreover, we hope that our reasoning behind each validation test can help other researchers determine what types of tests are appropriate for their particular research question. We are now in a time where computational tools are developed faster than our social scientific research can progress (DiMaggio, 2015). We believe it will be important for interested

---

[5] We would like to thank an anonymous reviewer for this suggestion.

social scientists to carefully consider the promises and limitations of these computational tools (Grimmer et al., 2022).

**Strengths, Limitations, and Future Research**

To our knowledge, our study is the first to examine LSS in long-term married couples across different types of conversations, expanding the extant literature on other linguistic synchrony (e.g., linguistic style matching) across interaction contexts (Bowen et al., 2017). The rich shared history between these couples provided a solid foundation to test our theoretical predictions.

According to the broaden-and-build theory of positive emotions (Fredrickson, 2013), the experience of positive emotions is associated with a widened cognitive space for people to retrieve a diverse set of past memories (Talarico et al., 2009; Wegner et al., 1985). In such a context, high LSS might therefore be an indicator of the lack of positive emotional experiences as conversation focuses on one topic instead of flowing from one to another. Yet, our research does not explicitly examine the "flow" of conversation, a question that future research can address (e.g., Yeomans et al., 2023). For instance, future research might employ structural topic modeling to quantitatively test the association between the topic distribution and people's emotional experiences (Roberts et al., 2014; also see Nakanishi et al., 2017). In addition to quantitative methods, qualitative methods such as discourse and conversation analysis might also be particularly useful in answering such a question (Kyprianou et al., 2015). It also remains to be seen whether our results are specific to married couples or generalize to other dyads with long-term, established relationships, such as long-term dating couples, close friends, or parent-child dyads.

Moreover, while we gave participants explicit instructions (i.e., to either talk about and try to resolve a conflict or talk about a topic of mutual enjoyment) as they initiated the conversations, some participants might hold different goals during the conversations. Whether and how these goals are represented by LSS and are associated with emotional experiences is an interesting avenue for future research to pursue[6]. It is possible that these goals serve as a mediator between LSS and emotional experiences (e.g., Gottman, 2017). For instance, low LSS might reflect a casual conversation, when topics can flow from one to another, suggesting a goal to connect, which could be associated with more positive emotional experiences in the pleasant conversation.

Other types of conversations should also be explored. An interesting type of conversation to consider is the sharing of good news (Gable & Reis, 2010). While this context should primarily be filled with positive emotions, research has shown that some listeners might respond in ways that diminish the sharers' positive emotional experiences, such as by being passive or destructive (Gable et al., 2004). In this case, LSS alone might not be sufficient in reflecting people's emotional experiences, and other methodological tools could be used to compliment the use of LSS, such as coding of the response styles of the listeners.

Additionally, it would be interesting to explore when two parties in a conversation endorse different goals. For instance, in interpersonal emotion regulation contexts, one person may want to feel more sadness (emotion up-regulation) while the other person wants to feel less sadness (emotion down-regulation) (Zaki, 2020). In such a situation, the psychological meanings of LSS might be multi-faceted or even ambiguous. We would argue for a multi-method approach combining LSS with other indices (e.g., self-report of goals, behavioral coding of regulation) to fully capture the emotional experiences in these conversations.

---

[6] We would like to thank an anonymous reviewer for this suggestion.

Future research might also consider using a more nuanced measure of emotional experiences. For instance, researchers could use a rating dial measure that continuously captures people's emotional experiences throughout the conversation. Instead of relying on post-conversation reports of emotional experiences (as we did here), a continuous rating dial could provide a fine-grained delineation of people's ups and downs as the conversation unfolds (Levenson & Ruef, 2007). It would also be interesting to examine whether emotional experiences flow along with pairwise LSS systematically within a conversation.

Finally, we recruited a sample from diverse socioeconomic and racial/ethnic backgrounds. Yet, our sample size—although comparable to other laboratory-based dyadic interaction studies (e.g., Babcock et al., 2014)—was only able to detect medium-sized effects larger than $r = .27$. In other words, while our sample was powerful enough to detect the effect size in the pleasant conversation, it may have been underpowered for the effect size in the conflict conversation. Future research is therefore needed to address this limitation.

**Conclusion**

This study examined the links between latent semantic similarity of natural conversations among married couples and their emotional experiences through a contextual lens. We tested predictions drawn from the shared-mind perspective and the broaden-and-build theory, and found supporting evidence for the role of conversation type as an important context moderating the link between latent semantic similarity and emotional experiences. Specifically, in pleasant conversations, higher semantic similarity was associated with lower positive emotional experiences. By contrast, in conflict conversation, semantic similarity showed a positive (albeit non-significant) association with emotional experiences. This work underscores the role of context in understanding the psychological meanings of latent semantic similarity and exemplifies how modern natural language processing tools can be used to evaluate theory-driven hypotheses in social psychology.

**Supplementary Information to "The More Similar, the Better? Association between Latent Semantic Similarity and Emotional Experiences Varied Across Conversation Types"**

In this document, we present the correlation table as well as the regression models when including covariates. These covariates include positive and negative emotional words, emotional style matching, linguistic style matching, number of turn pairs, and marriage satisfaction. We first present the models for pleasant conversations (Tables 1 to 7) and then for conflict conversations (Table 8 to 13).

**Table 1**

*Correlation matrix for the pleasant conversation*

|  | LSS | emotional experience | female | posemo | negemo | posemo_matching | negemo_matching | LSM | pair number |
|---|---|---|---|---|---|---|---|---|---|
| LSS |  |  |  |  |  |  |  |  |  |
| emotional experience | -0.25* |  |  |  |  |  |  |  |  |
| female | -0.072 | -0.034 |  |  |  |  |  |  |  |
| posemo | 0.084 | 0.06 | -0.24* |  |  |  |  |  |  |
| negemo | -0.19 | -0.038 | 0.16 | -0.21* |  |  |  |  |  |
| posemo_matching | 0.21* | -0.11 | -0.12 | 0.096 | 0.0087 |  |  |  |  |
| negemo_matching | -0.19 | 0.057 | 0.031 | -0.049 | 0.29** | -0.081 |  |  |  |
| LSM | 0.14 | 0.017 | -0.058 | 0.014 | 0.054 | 0.12 | 0.0041 |  |  |
| pair number | -0.21* | 0.11 | -0.0044 | -0.098 | 0.12 | 0.18 | -0.13 | 0.17 |  |
| marriage satisfaction | -0.15 | 0.32** | -0.027 | 0.22* | 0.065 | -0.2* | 0.12 | 0.046 | 0.061 |

*Note.* LSS = latent semantic similarity; posemo = positive emotional words; negemo = negative emotional words; LSM = linguistic style matching

**Table 2**

*Regression model controlling for emotional words in the pleasant conversation*

| term | b | SE | df | t | *p* |
|------|------|------|------|------|------|
| female | 0.07 | 0.15 | 46.4 | 0.471 | 0.64 |
| LSS | -0.32 | 0.13 | 81.6 | -2.41 | 0.018 |
| posemo | 0.03 | 0.08 | 87 | 0.347 | 0.729 |
| negemo | -0.1 | 0.11 | 83 | -0.913 | 0.364 |
| female x LSS | 0.15 | 0.16 | 42.4 | 0.953 | 0.346 |

*Note*. LSS = latent semantic similarity; posemo = positive emotional words; negemo = negative emotional words

**Table 3**

*Regression model controlling for emotional style matching in the pleasant conversation*

| term | b | SE | df | t | *p* |
|------|------|------|------|--------|-------|
| female | 0.05 | 0.14 | 45.4 | 0.368 | 0.715 |
| LSS | -0.29 | 0.14 | 78.2 | -2.12 | 0.038 |
| posemo_matching | -0.05 | 0.1 | 43.8 | -0.522 | 0.605 |
| negemo_matching | -0.01 | 0.1 | 47.2 | -0.0777 | 0.938 |
| female x LSS | 0.15 | 0.16 | 43 | 0.913 | 0.366 |

*Note.* LSS = latent semantic similarity; posemo = positive emotional words; negemo = negative emotional words

**Table 4**

*Regression model controlling for linguistic style matching in the pleasant conversation*

| term | b | SE | df | t | *p* |
|------|------|------|------|--------|-------|
| female | 0.05 | 0.14 | 45.5 | 0.38 | 0.706 |
| LSS | -0.31 | 0.13 | 81.5 | -2.3 | 0.024 |
| LSM | 0.05 | 0.11 | 45.6 | 0.473 | 0.638 |
| female x LSS | 0.15 | 0.16 | 43.1 | 0.917 | 0.364 |

*Note.* LSS = latent semantic similarity; LSM = linguistic style matching

**Table 5**

*Regression model controlling for turn pairs in the pleasant conversation*

| term | b | SE | df | t | *p* |
|------|------|------|------|--------|-------|
| female | 0.05 | 0.14 | 45.6 | 0.374 | 0.71 |
| LSS | -0.29 | 0.14 | 81 | -2.12 | 0.037 |
| pair number | 0.06 | 0.13 | 46.3 | 0.493 | 0.624 |
| female x LSS | 0.15 | 0.16 | 43.2 | 0.911 | 0.367 |

*Note.* LSS = latent semantic similarity

**Table 6**

*Regression model controlling for marriage satisfaction in the pleasant conversation*

| term | b | SE | df | t | *p* |
|------|-----|-----|------|--------|-------|
| female | -0.02 | 0.15 | 44.5 | -0.137 | 0.892 |
| LSS | -0.28 | 0.13 | 82 | -2.16 | 0.034 |
| marriage satisfaction | | | | | |
| | 0.08 | 0.12 | 83.5 | 0.655 | 0.514 |
| female x LSS | 0.16 | 0.16 | 40.7 | 0.988 | 0.329 |
| female x marriage satisfaction | | | | | |
| | 0.26 | 0.17 | 55.4 | 1.55 | 0.127 |

*Note.* LSS = latent semantic similarity

**Table 7**

*Regression model controlling for all covariates in the pleasant conversation*

| term | b | SE | df | t | *p* |
|------|-----|-----|------|---------|-------|
| female | -0.01 | 0.16 | 46.5 | -0.0739 | 0.941 |
| LSS | -0.29 | 0.15 | 70.6 | -2.02 | 0.047 |
| marriage satisfaction | | | | | |
| | 0.08 | 0.13 | 77.9 | 0.607 | 0.546 |
| pair number | 0.06 | 0.14 | 42.7 | 0.446 | 0.658 |
| LSM | 0.05 | 0.11 | 42 | 0.449 | 0.656 |
| posemo_matching | -0.02 | 0.1 | 42.4 | -0.199 | 0.843 |
| negemo_matching | 0.02 | 0.1 | 46.1 | 0.18 | 0.858 |
| posemo | -0.01 | 0.09 | 78.2 | -0.0961 | 0.924 |
| negemo | -0.11 | 0.11 | 73 | -0.973 | 0.334 |
| female x LSS | 0.17 | 0.17 | 40.3 | 1 | 0.322 |

female x marriage
satisfaction

| | | | | |
|---|---|---|---|---|
| 0.25 | 0.17 | 53.7 | 1.43 | 0.159 |

*Note.* LSS = latent semantic similarity; posemo = positive emotional words; negemo =
negative emotional words; LSM = linguistic style matching

**Table 8**

*Correlation matrix for the conflict conversation*

| | LSS | emotional experience | female | posemo | negemo | posemo_matching | negemo_matching | LSM | pair number |
|---|---|---|---|---|---|---|---|---|---|
| LSS | | | | | | | | | |
| emotional experience | 0.18 | | | | | | | | |
| female | 0.00048 | -0.12 | | | | | | | |
| posemo | 0.094 | -0.043 | 0.063 | | | | | | |
| negemo | 0.062 | -0.05 | 0.19 | -0.17 | | | | | |
| posemo_matching | 0.046 | 0.018 | 0.09 | 0.15 | 0.22* | | | | |
| negemo_matching | -0.24* | -0.28** | 0.14 | -0.027 | 0.44**** | 0.14 | | | |
| LSM | -0.027 | 0.18 | 0.067 | -0.15 | 0.14 | 0.0094 | 0.27** | | |
| pair number | -0.2* | -0.15 | 0.037 | -0.03 | 0.35*** | 0.072 | 0.38*** | 0.17 | |
| marriage satisfaction | -0.094 | 0.15 | 0.27** | -0.082 | 0.036 | 0.34*** | -0.046 | -0.2 | -0.067 |

*Note*. LSS = latent semantic similarity; posemo = positive emotional words; negemo = negative emotional words; LSM = linguistic style matching

**Table 9**

*Regression model controlling for emotional words in the conflict conversation*

| term | b | SE | df | t | *p* |
|---|---|---|---|---|---|
| female | 0.04 | 0.17 | 42.8 | 0.211 | 0.834 |
| LSS | 0.22 | 0.17 | 73.7 | 1.24 | 0.218 |
| posemo | -0.07 | 0.12 | 84.8 | -0.58 | 0.563 |
| negemo | -0.11 | 0.1 | 82.5 | -1.11 | 0.269 |
| female x LSS | 0.07 | 0.19 | 42.2 | 0.348 | 0.729 |

*Note*. LSS = latent semantic similarity; posemo = positive emotional words; negemo = negative emotional words

**Table 10**

*Regression model controlling for emotional style matching in the conflict conversation*

| term | b | SE | df | t | *p* |
|---|---|---|---|---|---|
| female | 0.04 | 0.17 | 43.4 | 0.222 | 0.825 |
| LSS | 0.11 | 0.17 | 73.1 | 0.665 | 0.508 |
| posemo_matching | 0.04 | 0.13 | 45.6 | 0.343 | 0.734 |
| negemo_matching | -0.28 | 0.14 | 46.9 | -2.03 | 0.048 |
| female x LSS | 0.1 | 0.19 | 42 | 0.497 | 0.622 |

*Note*. LSS = latent semantic similarity; posemo = positive emotional words; negemo = negative emotional words

**Table 11**

*Regression model controlling for linguistic style matching in the conflict conversation*

| term | b | SE | df | t | *p* |
|---|---|---|---|---|---|
| female | 0.04 | 0.17 | 43 | 0.201 | 0.841 |
| LSS | 0.2 | 0.17 | 75 | 1.2 | 0.233 |
| LSM | 0.24 | 0.15 | 81.3 | 1.59 | 0.115 |
| female x LSS | 0.07 | 0.19 | 41.9 | 0.383 | 0.704 |
| female x LSM | -0.09 | 0.17 | 47.7 | -0.51 | 0.612 |

*Note*. LSS = latent semantic similarity; LSM = linguistic style matching

**Table 12**

*Regression model controlling for turn pairs in the conflict conversation*

| term | b | SE | df | t | *p* |
|---|---|---|---|---|---|
| female | 0.03 | 0.17 | 44.1 | 0.189 | 0.851 |
| LSS | 0.16 | 0.17 | 73.9 | 0.937 | 0.352 |
| pair number | -0.12 | 0.12 | 45.1 | -1 | 0.321 |
| female x LSS | 0.08 | 0.19 | 42.8 | 0.435 | 0.666 |

*Note*. LSS = latent semantic similarity

**Table 13**

*Regression model controlling for marriage satisfaction in the conflict conversation*

| term | b | SE | df | t | p |
|---|---|---|---|---|---|
| female | 0.04 | 0.18 | 42.3 | 0.196 | 0.846 |
| LSS | 0.19 | 0.17 | 73 | 1.11 | 0.272 |
| marriage satisfaction | | | | | |
| | 0.06 | 0.16 | 77.9 | 0.392 | 0.696 |
| female x LSS | 0.11 | 0.2 | 40.2 | 0.567 | 0.574 |
| female x marriage satisfaction | | | | | |
| | 0.06 | 0.21 | 48.1 | 0.307 | 0.76 |

*Note*. LSS = latent semantic similarity

**Table 14**

*Regression model controlling for all covariates in the conflict conversation*

| term | b | SE | df | t | p |
|---|---|---|---|---|---|
| female | 0.06 | 0.18 | 40.1 | 0.349 | 0.729 |
| LSS | 0.11 | 0.18 | 66.6 | 0.603 | 0.549 |
| marriage satisfaction | | | | | |
| | 0.07 | 0.16 | 72.1 | 0.402 | 0.689 |
| pair number | -0.05 | 0.13 | 40.1 | -0.413 | 0.682 |
| LSM | 0.3 | 0.13 | 44.8 | 2.35 | 0.023 |
| posemo_matching | 0.05 | 0.14 | 47.8 | 0.343 | 0.733 |
| negemo_matching | -0.31 | 0.16 | 45.2 | -1.91 | 0.062 |
| posemo | -0.01 | 0.13 | 76.8 | -0.0666 | 0.947 |
| negemo | -0.04 | 0.12 | 76.7 | -0.329 | 0.743 |
| female x LSS | 0.14 | 0.2 | 38.5 | 0.688 | 0.496 |
| female x marriage satisfaction | | | | | |
| | 0.08 | 0.21 | 46.9 | 0.398 | 0.692 |

1    *Note*. LSS = latent semantic similarity; posemo = positive emotional words; negemo = negative

2    emotional words; LSM = linguistic style matchin